# Performance Examination of Symbolic Aggregate Approximation in IoT Applications

Suzana Veljanovska
Institute of Embedded Systems
ZHAW School of Engineering
Winterthur, Switzerland
veln@zhaw.ch

Hans Dermot Doran
Institute of Embedded Systems
ZHAW School of Engineering
Winterthur, Switzerland
donn@zhaw.ch

*Abstract*—**Symbolic Aggregate approXimation (SAX) is a common dimensionality reduction approach for time-series data which has been employed in a variety of domains, including classification and anomaly detection in time-series data. Domains also include shape recognition where the shape outline is converted into time-series data for instance epoch classification of archived arrowheads. In this paper we propose a dimensionality reduction and shape recognition approach based on the SAX algorithm, an application which requires responses on cost efficient, IoT-like, platforms. The challenge is largely dealing with the computational expense of the SAX algorithm in IoT-like applications, from simple time-series dimension reduction through shape recognition. The approach is based on lowering the dimensional space while capturing and preserving the most representative features of the shape. We present three scenarios of increasing computational complexity backing up our statements with measurement of performance characteristics.**

*Keywords*—**Symbolic Aggregate Approximation, Low Energy, IoT, Shape Recognition, Anomaly Detection**

## I. Introduction

### 1. Motivation

Automated Visual Inspection (AVI) is a common technique that uses computer vision to analyze images of products in the manufacturing process and detect defects and anomalies without human intervention [1], [2]. More precisely, it focuses on detecting anomalous shapes that deviate from the generally suitable silhouette. This plays major role in quality control and efficiency in the manufacturing process. For optimal energy and latency optimization, the AVI should perform directly on the edge.

Given that edge devices often have high resource constraints, it is crucial to design algorithms that are both optimized and computationally lightweight [3]. Another crucial aspect is the high costs of communication pointed out in [4] where despite using a low-power, wide-area, networking protocol like LoRaWAN for wireless communication consumes significantly more energy compared to the local processing performed on the edge device. This makes performing on-board data processing and abstraction for communication a viable approach. Therefore, it is most convenient to store and compute resources directly on the edge device, ideally at the same physical location as the data source.

Many proposed techniques specifically designed for AVI utilize machine learning algorithms and convolutional neural networks which are highly computationally intensive and not suitable for IoT platforms [5], [6], [7]. In general, when dealing with image processing tasks, the challenge becomes more significant due to their computationally intensive calculations. Our main aim is to achieve an effective shape classification while maintaining computational efficiency, allowing implementation on low-energy IoT platforms.

In the context of low-energy IoT applications, we chose to implement our algorithm on a Nordic (nRF5340) board [8] as representative of a commonly used low-power IoT semi-conductor device. We propose an approach for shape classification by reducing the images into time-series and down to strings utilizing SAX as an effective tool for mapping time-series into strings while preserving the crucial features of the signal. We then demonstrate the suitability of SAX by applying it to a temperature measurement time-series use case and extend it further to more complex task such as shape abstraction by incorporating two levels of abstraction to decrease the computational costs.

*2. Methodology*

The employed dimension reduction method decreases the dimensional space while efficiently retaining the key features of the shape. This approach relies on two abstractions, initially, transitioning from a 2D to a 1D space, followed by further reduction through conversion into a single string. As a final step, the strings are used for shape classification. The entire algorithm maintains low computational complexity which makes it compatible for deployment on IoT platforms.

II. THE DIMENSIONALITY REDUCTION WORKFLOW

The algorithm aims to reduce data dimensions by mapping 2D image into a string. It comprises two primary sequences depicted in Figure 1. Firstly, the image is converted into a time-series format and secondly the time-series is mapped with a single string.

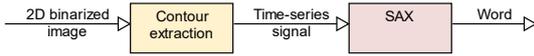

Figure 1. Dimensionality reduction flow

The algorithm requires a binarized image with a specific shape as input. The initial step involves converting the 2D image into a 1D time-series, accomplished through contour extraction of the shape. The values of the time-series represent the distance from the shape's center to the contour. Subsequently, the time-series is transformed into a symbol string (word) of a predefined length using the SAX algorithm [9].

*1. Time-series signal extraction from 2D images*

By employing the centroid distance function, we generate a shape time-series utilizing both, the centroid and the contour points to produce the time-series signal. Shape time-series, which are one-dimensional functions derived from the shape's contour, offer insights into its features. The basis of the time-series lies in the centroid distance function, which quantifies the distances from the contour points to the shape's centroid [10].

The initial step involves determining the centroid of the shape. Here, the shape area comprises the white pixels in the image. The shape itself is defined as the outline of the object. The centroid coordinates $(x_c, y_c)$ are calculated by averaging the pixel coordinates inside the shape, denoted as N (the number of pixels in the shape, i.e., the white pixels), and are computed using the relation (1) where $x_i$ and $y_i$ are the coordinates of the pixels contained in the shape.

$$\begin{cases} x_c = \frac{1}{N}\sum_{i=1}^{N} x_i \\ y_c = \frac{1}{N}\sum_{i=1}^{N} y_i \end{cases} \quad (1)$$

Once the centroid is obtained, the distance from the centroid to the point on the contour for each angle is determined as Euclidean distance between the two points using the relation (2).

$$r(\theta) = \sqrt{(x(\theta) - x_c)^2 + (y(\theta) - y_c)^2}, \; \theta = [0, 2\pi] \quad (2)$$

One representative point from the contour for each angle is considered. The main purpose of this sampling step is to get time-series signals of equal length allowing the shape comparison and classification later on since it is only reasonable to compare words of the same length.

Because the shape time-series relies solely on the centroid's position and the contour points, it remains unchanged when the shape is translated. However, this is not the case when the shape is rotated or scaled (expanded or compressed) [11].

*2. SAX algorithm*

The SAX algorithm requires three steps to convert the time-series into a string representation [12], [13].

Step 1. Data normalization step

Since it is meaningless to compare time-series with different offsets and amplitudes, the time-series needs to be normalized.

After the normalization step, the time-series still preserves its original shape. The normalization is performed utilizing the relation (3) where $\mu$ is the mean value and $\sigma$ (4) is the standard deviation of the signal.

$$X' = \frac{X - \mu}{\sigma} \quad (3)$$

$$\sigma = \sqrt{\frac{1}{N}\sum_{i=1}^{N}(x_i - \mu)^2} \quad (4)$$

Step 2. Dimensionality reduction via PAA

The Piecewise Aggregate Approximation (PAA) step reduces the signal to a desired length. A time-series $T$ of length $n$, $T = t_1, \ldots, t_n$ can be represented in a $w$-dimensional space by a $T' = t'_1, \ldots, t'_w$. Each $t_i$ element is calculated applying the relation (5)

$$t_i = \frac{w}{n}\sum_{j=\frac{n}{w}(i-1)+1}^{\frac{n}{w}(i)}(t_j) \quad (5)$$

The data is partitioned into $w$ segments of equal size, where $n$ represents the total number of signal samples. Within each segment, the mean value of the samples is assigned.

Step 3. Discretization step

Since SAX is a process which maps the PAA representation of the time-series into a sequence of letters, the last step of the algorithm is assigning a letter to each PAA segment. Each PAA segment gets assigned a symbolic letter. In addition to selecting the

PAA size, another parameter that is considered is the desired number of distinct letters to be represented in the word.

Let $a$ be the number of symbols $c_1,..., c_a$ which are used to discretize the time-series $\beta_1, \beta_2,..., \beta_{a-1}$ where $\beta_1 < \beta_2 < ... < \beta_{a-1}$ are the cuts on the Gaussian curve where each interval occupies equal part under the distribution curve. Each segment of $t_i$ will be coded as a symbol $x_i$ applying the equation (6).

$$x_i = \begin{cases} c_1, & t_i \leq \beta_1 \\ c_a, & t_i > \beta_{a-1} \\ c_k, & \beta_{k-1} < t_i \leq \beta_k \end{cases} \quad (6)$$

The $\beta$ interval cuts for the normalized signals are given with the look-up Table 1. According to the alphabet size, the $\beta$ intervals can be derived.

|   |   | a |   |   |   |   |   |
|---|---|---|---|---|---|---|---|
|   |   | 3 | 4 | 5 | 6 | 7 | 8 |
| β | $\beta_1$ | -0.43 | -0.67 | -0.84 | -0.97 | -1.07 | -1.15 |
|   | $\beta_2$ | 0.43 | 0 | -0.25 | -0.43 | -0.57 | -0.67 |
|   | $\beta_3$ |  | 0.67 | 0.25 | 0 | -0.18 | -0.32 |
|   | $\beta_4$ |  |  | 0.84 | 0.43 | 0.18 | 0 |
|   | $\beta_5$ |  |  |  | 0.97 | 0.57 | 0.32 |
|   | $\beta_6$ |  |  |  |  | 1.07 | 0.67 |
|   | $\beta_7$ |  |  |  |  |  | 1.15 |

Table 1. Look-up table for the β intervals

### 3. *Time-series use-case*

We demonstrate an application of time-series analysis in the context of temperature monitoring, for instance within buildings. Sensors are employed to gather temperature data within approximately one minute time intervals. Figure 2 (upper plot) illustrates the raw data collected over a two-week period. Now the question arises: how much of this data is relevant? How can we extract some relevant information from this extensive data, while minimizing computational resources?

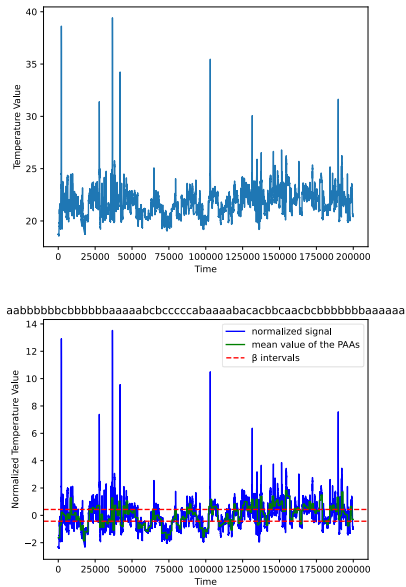

Figure 2. (Upper plot) Raw time-series. (Bottom plot) SAX time-series

We answer the question by utilizing SAX on the time-series shown on the bottom plot representing the signal as a sequence of letters. This approach offers a practical means to identify temperature variations. By analyzing the letters in the word, we can track each change of the letter and interpret it as a temperature shift instead of analyzing the whole time-series. Specifically, we have the flexibility to choose the alphabet size to determine whether we want to capture minor or major temperature changes. If we map the words with a larger alphabet, we retain sensitivity to small temperature fluctuations. Conversely, if we limit the alphabet length, as in our example, we focus on identifying larger temperature changes.

This logic extends to the selection of PAA segments size as well. Larger PAA segments sacrifice precision in temperature change detection that results into more efficient dimensionality reduction, while smaller segments provide greater precision at the cost of longer words.

### 4. *Shape recognition*

We extend the application of the SAX algorithm to shape recognition and classification. The SAX time-series of three shapes are depicted in Figure 3. The horizontal x-axis corresponds to the angles of the points on the contour, while the vertical y-axis represents the mapped pixel distances from the centroid to these points.

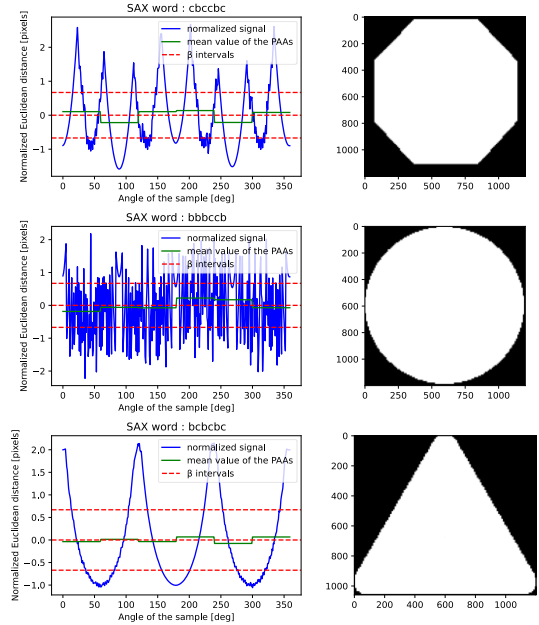

Figure 3. SAX time-series for shape recognition

Each shape is distinctly recognizable and exhibits unique characteristics. In the octagon and triangle shapes, there are eight and three peaks respectively in the amplitude, corresponding to the corners of the shapes. However, the time-series for the circle resembles noise due to signal normalization, which amplifies the small deviations. Nevertheless, this does not hinder shape recognition, as it is effectively distinguished from the other two shapes since the circle shape exhibits significantly higher frequency compared to the frequencies from the rest of the shapes.

## 5. Shape classification

The SAX algorithm can extract the unique characteristics of a shape, producing a SAX word as output. However, the algorithm itself is not rotation invariant, which means that a rotated shape is assigned a different SAX word in relation to the original (unrotated) one. When a signal undergoes rotational transformations, it leads to subtle shifts in the peaks of the signal and/or different values for these amplitude peaks. To be more precise, when the image is rotated along a single axis, it only causes the peaks to appear at different positions within the time-series. In our scenario, when the shape undergoes a 3D rotation along three axes, it leads to both, shift in the amplitude peaks as well as variation of the peak's values. This limitation represents an issue for shape classification applications, which should recognize image shapes without the rotation of the images affecting the result.

We address this limitation by generating a distinct set of words that captures variations due to rotation. In Figure 4, several images from the sets of rotations which are used to generate the words are displayed. This procedure is repeated for each shape to create a set of words, where each set encompasses unique words corresponding to various rotations of the shape. In addition, these sets are mutually exclusive, ensuring that there is no overlap of identical words across different shape sets.

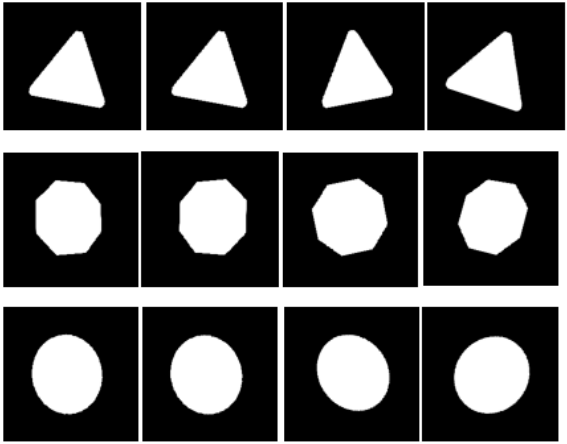

Figure 4. Samples of the rotated shapes sets

The final step is the classification process which involves brute-forcing through all word sets and calculating the distance between a candidate word and the remaining synthetic words. Algorithm 1 describes the distance computation when comparing two words. The comparison is conducted letter by letter, where the distance between identical or neighboring letters is zero. Otherwise, the distance gets assigned with the interval between the two letters. When the brute force search is done, the shape inherits the class of the word with the minimum distance.

**Algorithm: Distance between words computation**

```
1:     dist ← 0
2:     while (n < word_length) do
3:         letter_{1,n} ← c_i      → read the letter from the candidate word
4:         letter_{2,n} ← c_j      → read the letter for the synthetic word
5:         if i = j or abs(i - j) = 1 then  → check if the letters same or neighboring
6:             dist ← dist + 0
7:         else
8:             dist ← dist + (β_i … β_{j-1})    → add the intervals between letters
9:     return dist
10:    end procedure
```

Algorithm 1. Distance between words computation for classification

## III. RESULTS

Our primary aim was to implement the algorithm on an IoT hardware platform. To achieve this, as previously mentioned, we opted for a nRF5340 Nordic board due to its low power characteristics. Although the board features dual-core functionality, we exclusively utilized the application processor, operating at a frequency of 64 MHz, accompanied by 1MB of Flash memory and 512KB of RAM. The chosen operating system is Zephyr RTOS.

The SAX algorithm was deployed across three distinct hardware platforms, and the results are summarized in Table 2. It is evident that the execution time in C is significantly faster, approximately ($\sim 10^3$) times faster.

|  | Clock frequency | Execution time in C | Execution time Python | Difference |
|---|---|---|---|---|
| RaspberryPi4 (Cortex-A72 32-bits OS) | 1.8 GHz | 0.03200101s | 7.319624554s | 7.287623544s |
|  | 700 MHz | 0.08255687s | 19.07252964s | 18.98997277s |
|  | 300 MHz | 0.1638856s | 37.409953774s | 37.24565207s |
| PC (Intel(R) Core-TM i7-1065G7) | 1.3 GHz | 0.01340653s | 1.216410509s | 1.203003979s |
| nRF5340DK (ARM Cortex-M33) | 64 MHz | 0.09333026s | - | - |

Table 2. Performance results of SAX execution on three different HW platforms

## IV. CONCLUSIONS

Considering the computational demands of image processing, we utilize SAX as an effective tool for data reduction to transform time-series sequence into a single word that characterizes the key features of the sequence. Therefore, a conversion process to transform an image containing a shape to a time-series was designed, allowing us to implement the SAX algorithm. In such manner, by extracting the essential features from the time-series signal, SAX serves as a base for the shape classification. We address the issue of rotation sensitivity, which we integrate into our classification algorithm, by creating 3D rotated synthetic signs sets and apply a word-based comparison method for shape classification. We then validate the feasibility of our approach by integrating it into a low-power IoT platform, underscoring its potential for application in anomaly detection for factory automation.

ACKNOWLEDGMENT

I would like to thank Carlos Rafael Tordoya Taquichiri for providing the results in Table 2.